\documentclass[letterpaper, 10 pt, conference]{ieeeconf}  

\IEEEoverridecommandlockouts                              

\let\labelindent\relax

\usepackage{amsmath} 
\usepackage{amssymb}  
\usepackage{enumitem}
\usepackage{algorithm}
\usepackage[noend]{algpseudocode}
\usepackage{varwidth}
\usepackage{multirow}
\usepackage{booktabs}
\usepackage{siunitx}
\usepackage[normalem]{ulem}
\usepackage{graphicx}
\usepackage{lipsum}
\renewcommand{\baselinestretch}{0.934}

\usepackage[dvipsnames]{xcolor}

\newcommand{\resp}[1]{\textcolor{black}{#1}}
\DeclareMathAlphabet\mathbfcal{OMS}{cmsy}{b}{n}
\usepackage{geometry}
\title{\LARGE \bf
A preprocessing-based planning framework for utilizing contacts in high-precision insertion tasks
}

\author{Muhammad Suhail Saleem, Rishi Veerapaneni, and Maxim Likhachev
\thanks{All authors are with the Robotics Institute, Carnegie Mellon University, Pittsburgh, PA 15213, USA. {\small e-mail: \tt \{msaleem2, rveerapa, mlikhach\}@andrew.cmu.edu}}%
}

\begin{document}

\maketitle
\thispagestyle{empty}
\pagestyle{empty}

\begin{abstract}

In manipulation tasks like plug insertion or assembly that have low tolerance to errors in pose estimation (errors of the order of 2mm can cause task failure), the utilization of touch/contact modality can aid in accurately localizing the object of interest. Motivated by this, in this work we model high-precision insertion tasks as planning problems under pose uncertainty, where we effectively utilize the occurrence of contacts (or the lack thereof) as observations to reduce uncertainty and reliably complete the task. We present a preprocessing-based planning framework for high-precision insertion in repetitive and time-critical settings, where the set of initial pose distributions (identified by a perception system) is finite. The finite set allows us to enumerate the possible planning problems that can be encountered online and preprocess a database of policies. Due to the computational complexity of constructing this database, we propose a general experience-based POMDP solver, E-RTDP-Bel, that uses the solutions of similar planning problems as experience to speed up planning queries and use it to efficiently construct the database. We show that the developed algorithm speeds up database creation by over a factor of 100, making the process computationally tractable. We demonstrate the effectiveness of the proposed framework in a real-world plug insertion task in the presence of port position uncertainty and a pipe assembly task in simulation in the presence of pipe pose uncertainty.

\end{abstract}

\section{INTRODUCTION}

Manipulation tasks like plug insertion (Fig. 1) or assembly have low tolerance to pose estimation errors. \resp{Errors as low as 2mm can cause task failure\footnote{\resp{Tolerance numbers obtained from experimental data}}}. However, it is infeasible for perception to always perfectly localize the object of interest (which for plug insertion is the port for the plug). This is especially the case for semi-structured environments where the variations in working conditions like lighting changes, occlusions, and changes in the relative pose between the camera and the object, make perception challenging. 

A common strategy to counteract this problem is to use visual servoing \cite{visual_servoing}\cite{hill1979real}. Visual servoing is a technique where a robot is controlled using visual feedback from a sensor (typically) mounted on the robot’s end effector. A control loop determines the movement of the robot’s end effector such that an error function based on the visual input is minimized. However, these approaches are subject to challenges all perception systems are affected by, including varying lighting conditions and occlusions (created by the object in-hand or by obstacles in the environment)\cite{kragic2003visual}. In this work, we explore an alternate avenue by allowing the robot to use its contact/touch modality to accurately localize the object of interest and complete the high-precision insertion tasks, thereby circumventing the above-mentioned challenges.

Instead of requiring a single perfect pose estimate, we relax the perception system to estimate a set of hypothesis poses within which the object of interest lies. The manipulator then utilizes the occurrence of contacts (or the lack thereof) during execution as observations to exactly localize the object and thereby complete the task. We demonstrate that this binary contact observation is extremely powerful and can reliably localize the object at a resolution fine enough to complete high-precision insertion tasks. Conceptually, we frame this as a planning problem under pose uncertainty where the robot observes contacts during execution.

\begin{figure}
\centering
\includegraphics[width=0.75\linewidth]{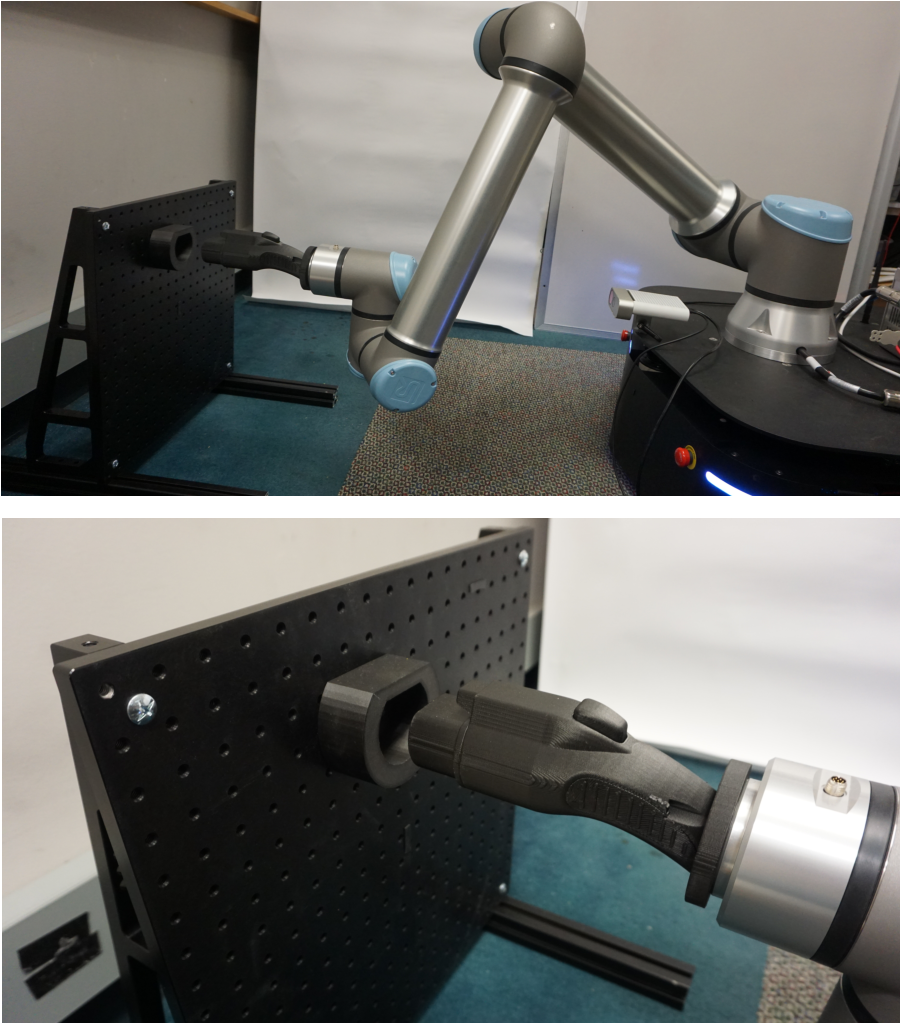}
\caption{Experimental setup for plug insertion using a UR10e manipulator.}
\label{fig:Real Robot} \vspace{-0.4cm}
\end{figure}

\resp{In this work, we focus on insertion tasks in semi-structured industrial settings where the set of initial pose distributions is finite (which in turn implies a finite set of possible planning problems)}. Prior works which have looked into semi-structured domains with finite problems have proposed enumerating the set of possible problems that can be encountered online and preprocessing a database of solutions \cite{CTMP}\cite{CTMP_Conveyor}. However, prior works solve deterministic problems, which only require the database to contain solution \textit{paths}. Due to pose uncertainty, our database needs to contain solutions for Partially Observable Markov Decision Processes (POMDPs) i.e. \textit{policies}. Naively constructing a database of such solutions is computationally infeasible due to the curses of dimensionality and history associated with solving POMDPs. Hence, it is imperative that intelligent choices are adopted to make the process of database creation scalable.

We exploit the fact that the problems in the database are similar to each other and propose a planning framework that iteratively constructs the database by using the solution of one problem as experience for solving the next. We propose a general experience-based POMDP solver, Experience-based RTDP-Bel (E-RTDP-Bel), that effectively utilizes prior experiences (solutions from similar planning problems) to speed up planning queries while maintaining strong bounds on solution quality and use it in our framework to efficiently construct the database of solution policies. 

The performance of the overall framework is demonstrated in the real world on the task of inserting a plug into a port using a UR10e manipulator. We utilize an ICP-based registration framework \cite{ICP1} to identify a discrete distribution of port poses. The results from the experiments highlight the robustness of the framework, succeeding 95\% of the runs. We also demonstrate the generality of the framework by using it on the task of pipe assembly in simulation. Performance analyses presented in Section \ref{Section: Results} show that the proposed E-RTDP-Bel algorithm improves planning times by over a factor of 100 while maintaining strong solution quality, making the database construction process feasible.

\resp{Succinctly, our contributions are:
\begin{enumerate}
    \item Developing a \textit{preprocessing-based planning framework} that solves high-precision insertion tasks under pose uncertainty for semi-structured settings.
    \item Developing a general \textit{experience-based RTDP-Bel algorithm} that uses experience from similar problems to significantly speed up planning queries while maintaining strong bounds on solution quality.
    \item Demonstrating the effectiveness of using binary contact information for accurate object localization.
\end{enumerate}}

\section{RELATED WORK}

We first discuss different frameworks used for solving high-precision insertion tasks. Then, we delve into  works that are closely related to our algorithmic contributions. 

\subsection{Framework}
 Visual servoing is a popular class of approaches used for solving high-precision insertion.
 Classical approaches in this field have explored three major ideas i) \textbf{Image-based visual servoing} which defines error metrics in the image space and computes end effector controls that minimizes the error \cite{IBVS1}, ii) \textbf{Position-based visual servoing} which continually estimates the pose of the object (using its model) and moves the end-effector towards a target pose \cite{PBVS1}, and iii) \textbf{Hybrid} approaches which intelligently combine the two ideas \cite{HBVS1}. Recent works have also trained deep neural network policies that map images to motor controls \cite{VS_learning}. In this work, we instead explore the use of an alternate sensing modality, tactile sensing, to close the feedback loop. 
 
Many prior works which incorporate tactile feedback focus on developing Particle Filter and Bayesian estimation methods to compute object poses that best explain the sequence of tactile observations made \cite{global_localization_via_touch}\cite{TBL_for_parts}\cite{saundPHD}. However, our work focuses more on active localization using tactile feedback, i.e. reasoning about the sequence of actions to take to quickly localize the object of interest. A popular idea here is to utilize a myopic framework that interleaves planning and execution \cite{TBL}\cite{TBL_for_parts}. In each planning cycle, actions are sampled and the action that maximizes an information gain metric is executed by the robot. This planning and execution cycle is repeated until the uncertainty is reduced sufficiently. As a consequence of being myopic and only reasoning about the next best action to take (as opposed to the sequence of actions to complete the task), these algorithms sacrifice solution quality which is critical in industrial settings. On the other hand, by formulating and solving the planning problem as a POMDP, we are able to achieve significantly better solution quality. This is evident from the performance comparisons with \cite{TBL} presented in Section \ref{Section: Results}.

Recently, reinforcement learning-based approaches for peg insertion have gained popularity \cite{PiH}. \cite{DRL} uses a long short-term memory model (LSTM) to estimate the Q value function to achieve peg-in-hole assembly. \cite{DDPG} proposes a model-driven DDPG algorithm to learn the general assembly policy for multiple peg-in-hole problems. We differentiate ourselves from prior literature in this field by posing the problem as an active localization task and explicitly reason about using contacts to localize the port/hole.

\subsection{Algorithm}
On an algorithmic level, we are inspired by \cite{E-Graphs} which introduces the notion of Experience Graphs for deterministic settings. Their approach uses solutions from similar episodes to construct a graph that represents the underlying connectivity of the space required to complete the task. Given a new planning query, the algorithm attempts to reuse this graph as much as possible by constructing a heuristic function that makes the search prioritize exploring regions around prior experiences. In this work, we extend the idea of Experience Graphs to POMDPs. A variety of algorithms have been proposed to solve POMDPs. RTDP-Bel \cite{RTDP-Bel} is a popular heuristic-search-based algorithm that exhibits great sample efficiency and provides strong solution guarantees. We augment RTDP-Bel to reuse prior experiences to significantly speed up planning while continuing to maintain solution guarantees. 


The idea of preprocessing a database of solutions to eliminate or speed up planning online has existed in different forms over the years \cite{CTMP}\cite{SubgoalGraphs}\cite{PRM}. Due to the computational complexity of solving POMDPs and the need for strong online performance, our approach utilizes a similar idea. 

\section{PRELIMINARY}

POMDPs are used to model sequential decision-making problems in non-deterministic settings where the state of the system is not directly observable. Instead, indirect observations are used to infer the system state. A discrete-time goal-POMDP can be characterized by the problem tuple $P = \textlangle \mathcal{S}, \mathcal{A}, \mathcal{Z}, \mathcal{T}, \mathcal{O}, \mathcal{C}, \mathcal{G} \textrangle $.

At every timestep, the system is at an unknown state $s \in S$ and executes an action $a \in A$. This transitions the system to a new state $s’$ dictated by the stochastic transition model $\mathcal{T}$. 
\begin{equation}
    \mathcal{T}(s, a, s') = P(s' | s, a), \ \forall \ s, s' \in \mathcal{S}, \ a \in \mathcal{A}
\end{equation}
After every transition, a noisy observation $z \in \mathcal{Z}$ is made, represented by the observation model $\mathcal{O}$.
\begin{equation}
   \mathcal{O}(s, a, z) = P(z | s, a), \ \forall z \in \mathcal{Z},\ s \in \mathcal{S},\ a \in \mathcal{A}
\end{equation}


As the state of the system is not directly observable, the transition and observation models are used to maintain a belief of the system state at every timestep. The belief state is a probability distribution over the possible system states and is a Markovian state signal that succinctly represents the history of actions taken and observations made. After executing action $a$ and receiving observation $z$ at timestep $t+1$, the belief state can be updated as follows:

\begin{equation}
\begin{gathered}
    b_{t+1}(s') =  b_a^z(s') = \sum_{s \in \mathcal{S}} b_t(s)\mathcal{T}(s, a, s') \mathcal{O}(s, a, z)
\end{gathered}
\end{equation}

Given a belief state and an action, the probability of transitioning to successor beliefs can be computed by combining the transition and observation probabilities. This essentially makes the problem a completely observable MDP in the belief space. $\mathcal{C}(b, a)$ is the cost incurred for executing action $a$ from belief state $b$.
The shortest path problem for POMDPs can then be stated as computing a policy $\pi$ in the belief space (mapping from belief states to actions), that takes the system from a start belief $b_{start}$ to any belief state in the goal set $\mathcal{G}$ while minimizing the expected cost incurred.

\section{PROBLEM FORMULATION}

In this section, we formulate the active localization using contacts problem as a POMDP.  Given a robot manipulator, let $\mathcal{R}$ represent its state space and $\mathcal{A}$ the discrete action space. The observation space $\mathcal{Z}$ comprises
\begin{enumerate}
    \item The robot's state (encoder readings), and
    \item A binary flag indicating the occurrence of contacts/collisions (F/T sensor readings or joint torques)
\end{enumerate}
Let $H_{start}$ represent the discrete distribution of hypothesis poses of the object of interest identified by a perception system. For ease of explanation and discussion, we assume this distribution is uniform (our approach can be extended to non-uniform distributions with minor modifications). This allows us to represent $H_{start}$ as a discrete set of hypothesis poses $H_{start} = \{h_1, h_2, ...\ h_n : h_i \in SE(3) \}$. 

\resp{
Unlike typical robotics problems, in our case, the stochasticity in transitions and observations arises as a result of uncertainty in the model of the environment i.e. the object pose. We assume the dynamics and the observations to be perfect given an object pose $h_i$. Meaning, $P(r'|r, a, h_i)$ and $P(z|r, a, h_i)$ are all either zero or one. The prevalence of high-quality manipulators in industries and the very simple and reliable observation space (contacts) make the perfect dynamics and observations assumption practical. Our real-world results detailed in Section \ref{Section: Results} are a testament to this. We also assume that the environment is static, i.e., robot actions do not change the state of the object being localized.
}

Hence, the state of the system $s$ for our problem contains both the robot state and the environment state, i.e. the object pose. This is represented by the tuple $(r, h)$. If $s$ = $(r, h)$ and $s'$ = $(r', h')$, the transition and observation models for our system state can be defined as follows.
\begin{equation}
\begin{gathered}
    \mathcal{T}(s, a, s') = P(s'|s, a) = \begin{cases} 
            1 \hspace{0.3cm} \ h' = h, \ P(r' | r, a, h) = 1 \\
            0 \hspace{0.3cm} \text{Otherwise}
        \end{cases} \\
    \mathcal{O}(s, a, z) = P(z | s, a) = \begin{cases} 
        1 \hspace{0.3cm} \ P(z| r, a, h) = 1 \\
        0 \hspace{0.3cm} \text{Otherwise}
        \end{cases} \\
\end{gathered} \vspace{-0.1cm}
\end{equation}
\resp{Essentially the transition model describes that executing an action $a$ from a fully observed state $s = (r, h)$ will deterministically transition the robot to a new state $r'$ while leaving the environment state (i.e. the object pose) unchanged.}  

The belief state is a probability distribution over the state space. Under the uniform initial pose distribution and perfect observation assumptions, the belief state can be represented by a tuple $(r, H)$, where $r$ is the robot state and $H = \{h_1, h_2 .. \}$ is the discrete set of hypothesis poses. As the task is to localize the object, the set of goal beliefs $\mathcal{G}$ is given by belief states that contain exactly one hypothesis pose. 
\begin{figure}
\centering
\includegraphics[width=0.95\linewidth]{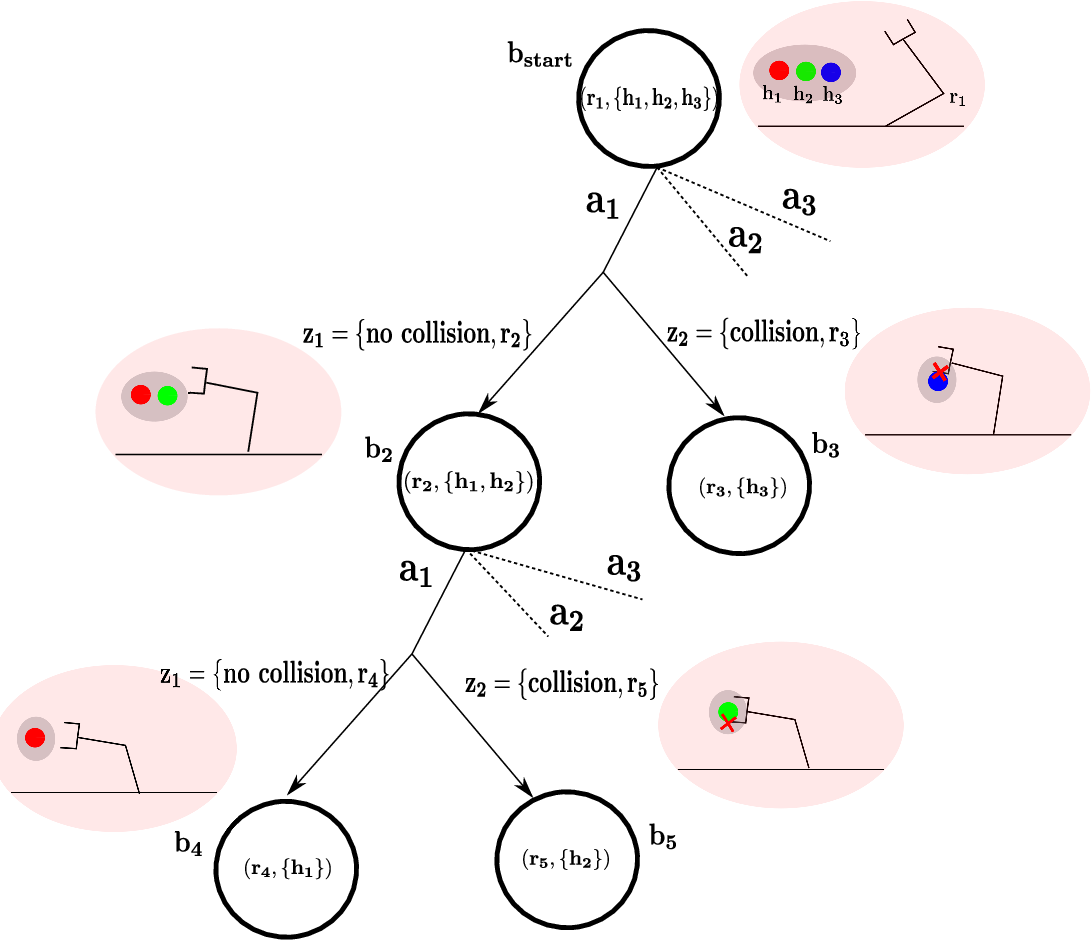}
\caption{Belief tree for the active localization using contacts problem. The robot uses the occurrence of contacts (or the lack thereof) to reduce uncertainty, i.e., the hypothesis pose set. }
\label{fig:belief tree} \vspace{-0.4cm}
\end{figure}

Fig \ref{fig:belief tree}. represents a portion of the belief tree. Executing an action from a belief state that results in different observations under different hypotheses leads to successor beliefs with reduced uncertainty. As can be seen in the figure, executing action $a_1$ from belief state $b_2$ which contains two hypothesis poses $\{h_1, h_2\}$, results in a collision observation under $h_2$ and uninterrupted execution under $h_1$. Executing this action allows us to disambiguate between the hypotheses $h_1$ and $h_2$ thereby localizing the object.


The planning problem representing the active localization task can therefore be stated as \textit{given a robot start state $r_{start}$ and the set of initial hypothesis poses $H_{start}$, compute a belief space policy that reduces the hypothesis set to a single element while minimizing the expected distance traveled.}

The assumptions we make about the above-formulated active localization problem can be concisely listed as follows: 
\begin{itemize}
    \item[\textbf{A1}] Realizable setting, i.e., the groundtruth object pose is contained in the initial hypothesis set $H_{start}$.
    \item[\textbf{A2}] Deterministic transitions and perfect observations given the environment state, i.e., object pose. 
    \item[\textbf{A3}] Known geometric model of the object of interest (used to compute expected observations).
    \item[\textbf{A4}] Static environment, i.e., the object of interest remains static upon interaction.
\end{itemize}

\resp{It should be emphasized that our approach can be extended to incorporate noisy transition and observation models with minor adjustments. However, in such cases, we can anticipate a decrease in planning speed due to the expanded search space. The assumption of a realizable setting ensures that the observations obtained from real-world interactions align with at least one of the hypotheses proposed in $H_{start}$. If none of the hypotheses can explain the observations, all of them will be eliminated, indicating that the task is unachievable.}

\section{APPROACH}

We are interested in semi-structured domains where the set of localization planning problems that can be encountered is finite and known ahead of time. This is defined by a fixed robot start state $r_{start}$ and the finite set of initial distributions that can be encountered $\mathcal{H} = \{H_{start}^1, H_{start}^2 ... \}$. As the set of problems is known, we propose to preprocess solutions to the problems ahead of time and construct a database of policies $\Pi = \{ \pi_1, \pi_2 ... \}$ that localize the object of interest for each possible distribution $H_{start}^i$. Online, based on the problem at hand, the solution policy from the database is retrieved and executed. 

Constructing this database naively by solving each problem independently is computationally expensive. Based on the key observation that the problems in the database are similar to each other, we propose to iteratively construct the database by using the solution of one problem as experience for solving the next. To this end, we develop a general experience-based POMDP solver, E-RTDP-Bel, that uses solutions of similar problems to speed up planning queries and use it in our problem domain. 

We start off by describing RTDP-Bel \cite{RTDP-Bel}, the POMDP solver that forms the basis of our planning framework. Followed by which we describe E-RTDP-Bel and move on to describing how it is used in the context of our domain. 

\subsection{RTDP-Bel}
RTDP-Bel \cite{RTDP-Bel} is a direct adaptation of RTDP \cite{RTDP} to POMDPs. Based on asynchronous value iteration, it converges to the optimal value function and policy for only the relevant states in the belief space. Outlined in Alg \ref{Alg1: RTDP-Bel}, it operates by performing a series of greedy rollouts/searches. Each rollout in RTDP-Bel starts from the initial belief state $b_{start}$ (Line \ref{Alg1:Start}) and terminates when a belief state in the goal set $\mathcal{G}$ is reached (Line \ref{Alg1: Goal}). Each iteration of the rollout consists of sampling a state from the current belief, evaluating the effect of executing different actions, selecting the best one, and updating the value estimate accordingly (Lines \ref{Alg1:Q-estimates} - \ref{Alg1: Update Value}). Then, the next state and observation are sampled and the new belief is computed (Lines \ref{Alg1: Sample Next State} - \ref{Alg1:New belief}).

\begin{algorithm}
\begin{small}
\caption{\textsc{RTDP-Bel}}
\label{Alg1: RTDP-Bel}
\begin{algorithmic}[1]
\While {\textsc{Not Converged}}
\State $b = b_{start}$ \label{Alg1:Start}
\State \textbf{Sample} state $s$ with probability $b(s)$
\While {$b \notin \mathcal{G}$}\label{Alg1: Goal}
\State \begin{varwidth}[t]{\linewidth} \label{Alg1:Q-estimates}
    \textbf{Evaluate} the value of executing each action $a \in \mathcal{A}$ \par from 
    belief state $b$ as: \vspace{-0.1cm}
    \begin{equation*}\vspace{-0.1cm}
    Q(b, a) = \mathcal{C}(b, a) + \sum_{z \in \mathcal{Z}} P(z | b, a) V(b_a^z)
    \end{equation*} 
    \Comment{\textbf{Use $\mathbf{V(b_a^z) = heur(b_a^z)}$ if not initialized}}
  \end{varwidth}
\State \textbf{Select} action $a_{best}$ that minimizes $Q(b, a)$
\State \label{Alg1: Update Value} \textbf{Update} value of belief state $V(b) = Q(b, a_{best})$
\State \textbf{Sample} \label{Alg1: Sample Next State} next state $s'$ with probability $\mathcal{T}(s, a_{best}, s')$
\State \textbf{Sample} observation $z$ with probability $\mathcal{O}(s, a_{best}, z)$
\State \textbf{Compute} \label{Alg1:New belief} $b_a^z$ and set $b := b_a^z \text{ and } s:= s'$
\EndWhile
\EndWhile
\end{algorithmic} 
\end{small} 
\end{algorithm}
The heuristic function $\mathbf{heur}$ used to initialize the value estimates (Line \ref{Alg1:Q-estimates}) is critical to the performance of the algorithm. An informed heuristic function can speed up convergence by encouraging the search to explore only relevant portions of the belief space. \cite{POMHDP} shows that if the heuristic function is an underestimate of the optimal value function, then inflating the heuristic values by $\varepsilon$, allows each search iteration to be more strongly guided by the heuristic, converging to a $\varepsilon$-suboptimal solution significantly quicker.

\subsection{E-RTDP-Bel: Experience-Based RTDP-Bel}


\begin{figure*}
\centering
  \includegraphics[width=0.9\textwidth]{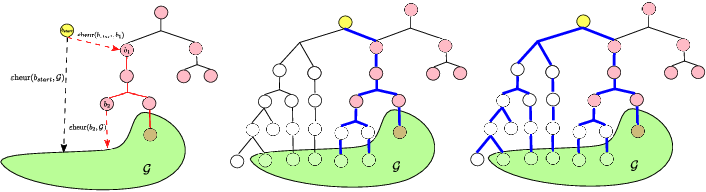}
  \caption{left) $b_{start}$, $\mathcal{B}^E$, and $\mathcal{G}$ are highlighted in yellow, maroon, and green respectively. The instantaneous transitions/jumps relevant for computing $\text{heur}^E$ for $b_{start}$ are indicated with dashed arrows. Under a sufficiently high $\varepsilon$, the experience heuristic prefers the policy highlighted in red, encouraging the search to move towards $\mathcal{B}^E$, and the value of this policy is used as $\text{heur}^E(b_{start})$. middle and right) Contrast the region of the belief space $\mathcal{B}$ explored by RTDP-Bel to converge to a solution (highlighted in blue) when using $\text{heur}^E$ (under high $\varepsilon$) vs $\text{heur}$, respectively. 
  }\label{Fig: E-Graph}
\end{figure*}

\subsubsection{Idea} The intuitive idea behind Experience-Based RTDP-Bel is that \textit{similar planning problems likely have similar solutions}. Hence, given a problem and the solution to a similar problem, exploring regions of the belief space around the solution is likely to solve the current problem.

Different from the original belief MDP of the planning problem $\mathcal{B}$, the approach explicitly maintains an experience MDP $\mathcal{B}^E$ which is a significantly smaller subset of the original belief space, i.e. $\mathcal{B}^E \subset \mathcal{B}$. $\mathcal{B}^E$ is constructed from solutions of previous planning problems and at its essence represents the portion of the belief space relevant to the planning problems. A simple technique to construct $\mathcal{B}^E$ would be to take the union of all transitions in previous solution policies. The key idea behind our approach then is to speed up planning queries by reusing portions of solution policies from previous problems as much as possible, i.e., portions of $\mathcal{B}^E$. This reduces the need for searching large areas of the original belief space $\mathcal{B}$.

We achieve this by constructing an intelligent heuristic function, the experience heuristic ($\text{heur}^E$), using the original heuristic defined for the problem domain ($\text{heur}$), the goal set for the problem $\mathcal{G}$, and the experience MDP $\mathcal{B}^E$. The experience heuristic biases the search iterations in RTDP-Bel towards $\mathcal{B}^E$ when following parts of $\mathcal{B}^E$ is likely to get the search closer to the goal. We use $\text{heur}^E(b)$ as shorthand for $\text{heur}^E(b, \mathcal{G})$ and is the heuristic estimate of the expected cost to reach the goal set of the problem from the belief $b$,

\begin{equation}\label{eqn: exp_heuristic}
    \begin{gathered}
      \text{heur}^E(b) = \min 
        \begin{cases} 
            {\displaystyle \varepsilon \text{heur}(b, b') + \text{heur}^E(b') \hspace{1cm} \forall \ b' \in \mathcal{B}}  \vspace{0.2cm}
              \\ 
            {\displaystyle \mathcal{C}(b, a) \ + \sum_{z \in \mathcal{Z}} P(z|a, b)\ \text{heur}^E(b^z_a)} \vspace{-0.2cm}\\
                \hspace{3.8cm} {\displaystyle \forall \ (b, a) \in \mathcal{B}^E}
                          \end{cases}
    \end{gathered}
\end{equation}

Essentially, given a belief state $b$, the experience heuristic computes the value of the optimal policy from $b$ to $\mathcal{G}$ that is composed of two kinds of transitions, i) Instantaneous jumps that take the system between any two belief states in the belief space $\mathcal{B}$. The cost of these instantaneous transitions is equal to the heuristic value inflated by the penalty term $\varepsilon$. ii) Transitions that are part of the belief MDP $\mathcal{B}^E$ which incur their true cost. By penalizing the (pseudo) transitions that lie outside the experience MDP, the heuristic encourages the search to reuse as much of $\mathcal{B}^E$ as possible. As the penalty $\varepsilon$ increases, the search will go farther out of its way to reuse experiences. We run RTDP-Bel with this newly defined experience heuristic to speed up planning. We refer to this approach as Experience-based RTDP-Bel (E-RTDP-Bel). Fig \ref{Fig: E-Graph}. provides a visual depiction of the experience heuristic idea.

\subsubsection{Implementation} 

Computing the experience heuristic as defined in Eqn \ref{eqn: exp_heuristic}. is computationally intractable. Given a belief state $b$, computing $\text{heur}^E$ involves reasoning over the instantaneous jumps from $b$ to every state in $\mathcal{B}$. However, if the original heuristic follows the $h$-consistency property, i.e., $\text{heur}(b_1, b_3) \leq \text{heur}(b_1, b_2) + \text{heur}(b_2, b_3), \forall \ b_1, b_2, b_3 \in \mathcal{B}$, then $\varepsilon \text{heur}(b) \leq \varepsilon \text{heur}(b, b') + \varepsilon \text{heur}(b') \ \forall \ b' \ \notin \mathcal{B}^E \cup \mathcal{G}$. Therefore, for all $b \in \mathcal{B}$, the instantaneous actions can be restricted to a significantly smaller subset, $b' \in \mathcal{B}^E \cup \mathcal{G}$. Now, instead of reasoning over instantaneous transitions to every state in the belief space, we only need to reason about jumps to the experience MDP $\mathcal{B}^E$ or the goal set $\mathcal{G}$.


\begin{equation}\label{eqn: exp_heuristic_2}
    \begin{gathered}
      \text{heur}^E(b) = \min 
        \begin{cases} 
            {\displaystyle \varepsilon \text{heur}(b, b') + \text{heur}^E(b') \hspace{0.38cm} \forall \ b' \in \mathbfcal{B}^E \cup \mathbfcal{G}}  \vspace{0.2cm}
              \\ 
            {\displaystyle \mathcal{C}(b, a) \ + \sum_{z \in \mathcal{Z}} P(z|a, b)\ \text{heur}^E(b^z_a)} \vspace{-0.2cm}\\
                \hspace{4cm} {\displaystyle \forall \ (b, a) \in \mathcal{B}^E}
                          \end{cases}
    \end{gathered}
\end{equation}

Computing the experience heuristic for every state encountered in the search is still non-trivial. However, computing $\text{heur}^E$ for any state $b$ encountered in RTDP-Bel can be reduced to a simple linear pass over the states in $\mathcal{B}^E \cup \mathcal{G}$ if the experience heuristic for the states in $\mathcal{B}^E$ is known.
\begin{equation}\label{eqn: exp_heuristic_3}
      \displaystyle {\text{heur}^E(b) = \min_{b' \in \mathcal{B}^E \cup \mathcal{G}} \varepsilon \text{heur}(b, b') + \text{heur}^E(b') \ \ \ \forall \ b \notin \mathcal{B}^E \cup \mathcal{G}}
\end{equation}

As the experience heuristic for states in $\mathcal{B}^E$ is only dependent on $\mathcal{G}$ and $\mathcal{B}^E$ itself, prior to running RTDP-Bel we precompute $\text{heur}^E$ for all states in $\mathcal{B}^E$ by performing a series of asynchronous Bellman updates (outlined in Alg 2).



\begin{algorithm}
\begin{small}
\caption{\textsc{Experience Heuristic}}
\begin{algorithmic}[1]
\Procedure {\textsc{Precompute Values}}{$\mathcal{B}^E, \mathcal{G}, \text{heur}$}
\For{$b \in \mathcal{B}^E$}
\State $\text{heur}^E(b) = \varepsilon \text{heur}(b)$ \Comment{initialization}
\EndFor
\While {\textsc{Not Converged}}
\For{$b \in \mathcal{B}^E$}
\State Compute $\text{heur}^E(b)$ as defined in Eqn \ref{eqn: exp_heuristic_2}.
\EndFor
\EndWhile
\EndProcedure
\end{algorithmic}
\label{Alg2}
\end{small} 
\end{algorithm}

For any belief state $b$, $\text{heur}^E(b)$ is upper bounded by $\varepsilon \text{heur}(b)$ (as $\text{heur}^E(b') = 0$, if $b' \in \mathcal{G}$). Hence, if $\text{heur}$ is admissible, i.e., underestimates the cost of transitions, the solution is guaranteed to be $\varepsilon$-suboptimal. This property directly follows from \cite{POMHDP}, which proved that a solution policy found by RTDP-Bel using an $\varepsilon$-inflated admissible heuristic is guaranteed to be $\varepsilon$-suboptimal.

\subsection{Using E-RTDP-Bel For Database Preprocessing}\label{subsection: Using E-RTDP-Bel for Preprocessing}


\begin{figure}
\centering
\includegraphics[width=\linewidth]{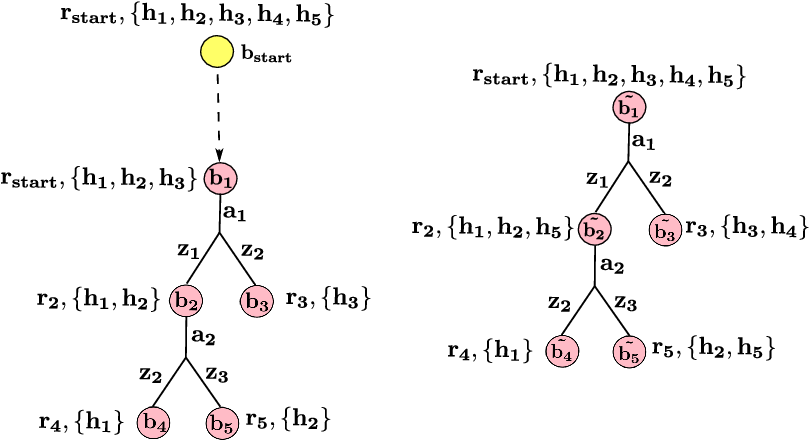}
\caption{left) Getting from the start belief of a more complex problem $p'$ (yellow) to the solution of a simpler problem $p$, i.e., $\mathcal{B}^E_{naive}$ (maroon) can be non-trivial. right) Rolling out the solution of the $p$ from $p'_{start}$ makes the experience directly accessible from $p'_{start}$.}
\label{fig:E_Graph_rollout} \vspace{-0.4cm}
\end{figure}

In this subsection, we show how E-RTDP-Bel is 
used to speed up constructing the database of policies $\Pi$ (Alg \ref{Alg:Preprocessing framework}).


Given the set of localization problems defined by the robot start state $r_{start}$ and the set of initial object pose distributions $\mathcal{H} = \{H_{start}^1, H_{start}^2 ... \}$, we can maximize the benefit of E-RTDP-Bel by solving the problems in an intelligent order. The closer the experience MDP $\mathcal{B}^E$ is to the solution policy of the problem being solved, the more speed up we are likely to observe, as larger portions of $\mathcal{B}^E$ can be reused. Hence, we choose to solve the problems in increasing order of uncertainty as it will allow us to reuse solution policies from simpler problems to construct rich $\mathcal{B}^E$'s which more closely resemble solutions to complex problems.

A subtle but important observation is that the start beliefs between the problems can be very different due to the initial uncertainty $H_{start}^i$ being different. For example, let the start belief of a simple problem $p$ be $p_{start} = (r_{start}, \{h_1, h_2, h_3\})$ and the start belief of a more complex problem $p'$ be $p'_{start} = (r_{start}, \{h_1, h_2, h_3, h_4, h_5\})$. In this case, directly using the solution policy of $p$ as the experience MDP for solving $p'$ might not be beneficial as getting from the start state of $p'$ to the experience is non-trivial in the belief space. We refer to the experience MDP which directly uses the solution of a similar problem as $\mathcal{B}^E_{naive}$. Hence, in this case getting from $p'_{start}$ to $\mathcal{B}^E_{naive}$ involves eliminating at least 2 hypothesis poses ($\{h_4, h_5\}$), and identifying actions that can achieve this is non-trivial. As a result, the impact of the experience is reduced.

To counter this issue, we transform $\mathcal{B}^E_{naive}$ in the belief space so it is easily accessible from $p'_{start}$. We achieve this by rolling out the policy of $p$ from $p'_{start}$ (as opposed to $p_{start}$ itself) and using the resulting MDP as the experience MDP $\mathcal{B}^E$. To roll out the solution policy computed for $p$ from $p’_{start}$, we represent the solution policy as a mapping from histories to actions (as opposed to belief states to actions). Here, history corresponds to the sequence of actions taken and observations made. For example, in Fig \ref{fig:E_Graph_rollout}, the belief state $b_2$ can be abstracted out as the state reached by executing action $a_1$ and receiving observation $z_1$. 
In the modified representation of the policy, history $\{a_1, z_1\}$ is mapped to action $a_2$. Hence, action $a_2$ is applied from the history $\{a_1, z_1\}$ in the more complex problem as well (which corresponds to belief state $\Tilde{b_2}$). Constructing $\mathcal{B}^E$ through this rollout procedure ensures that the experience can be reused directly, maximizing the speedup. This is evident from the ablation studies presented in Section \ref{Section: Results}.

\begin{algorithm}
\begin{small}
\caption{\textsc{Preprocessing Framework}}\label{Alg:Preprocessing framework}
\begin{algorithmic}[1]
\Procedure{\textsc{Preprocess}}{$r_{start}, \mathcal{H}$}
\State $\Pi = \emptyset$
\State Order $\mathcal{H}$ in increasing order of uncertainty
\For{$H_{start}^i \in \mathcal{H}$}
\State $b_{start} = (r_{start}, H_{start}^i)$
\State Pick $\pi^j$ from $\Pi$ with most similar start state uncertainty
\State $\mathcal{B}^E = \text{Rollout } \pi^j \text{ from } b_{start}$
\State Use Alg \ref{Alg2}. to precompute $\text{heur}^E(b) \ \forall \ b \in \mathcal{B}^E$ \State Use Eqn \ref{eqn: exp_heuristic_3}. to compute $\text{heur}^E(b) \ \forall \ b \notin \mathcal{B}^E \cup \mathcal{G}^i$
\State \parbox[t]{203pt}{Compute solution policy $\pi^i$ from $b_{start}$ to $\mathcal{G}^i$ using RTDP-Bel with $\text{heur}^E$\strut \Comment{E-RTDP-Bel}} 
\State $\Pi.\textsc{Insert}(\pi^i)$

\EndFor
\EndProcedure
\end{algorithmic}
\end{small} 
\end{algorithm}\vspace{-0.2cm}



\section{RESULTS}\label{Section: Results}
We evaluate the performance of our framework on i) A real-world plug insertion task incorporating 3D positional uncertainty of the plug port, and ii) A simulated pipe assembly task incorporating 4D pose uncertainty of the pipes. 

\subsection{Plug Insertion Task}

\subsubsection{Setup}
The task is to insert a plug into its port using a UR10e manipulator. The position of the port is constrained to be within a bounded volume (2m x 2m x 2m) and is allowed to yaw within a range of -15 to +15 degrees (roll, pitch are fixed). As 2mm errors in position estimation can cause failures, the task is modeled under 3D positional uncertainty.

 \resp{We use a simple perception module rooted in ICP (Iterative Closest Point) \cite{ICP1} to identify a discrete distribution of hypotheses port poses $H_{i}$. Given a scene, we capture a point cloud from the static camera (Microsoft Azure Kinect DK) and use ICP (with the model of the port) to estimate the port pose. This pose estimate is converted into a discrete distribution using a simple rule-based approach. We represent the 3D positional uncertainty as a cuboid centered around the pose estimate. This is then discretized at a resolution of 2mm to create the hypotheses set $H_i$. The rules for generating the cuboid of uncertainty  given the pose estimate are constructed by comparing the groundtruth poses with the ICP (noisy) estimates at different regions in the workspace. 
 Given this rule-based uncertainty model and the workspace, we query the model at all (discretized) poses within the workspace to create the set of possible initial distributions $\mathcal{H} = \{H_1, H_2 … \}$. The set contains initial uncertainties ranging from 4mm to 3cm along each axis, meaning that $H_i$ contains up to $15^3$ poses. The uncertainty model is not the focus of our framework, more sophisticated approaches can be used with our planning framework.}

\subsubsection{Impact of Reusing Experiences}
\begin{figure*}
\centering
  \includegraphics[width=0.9\textwidth]{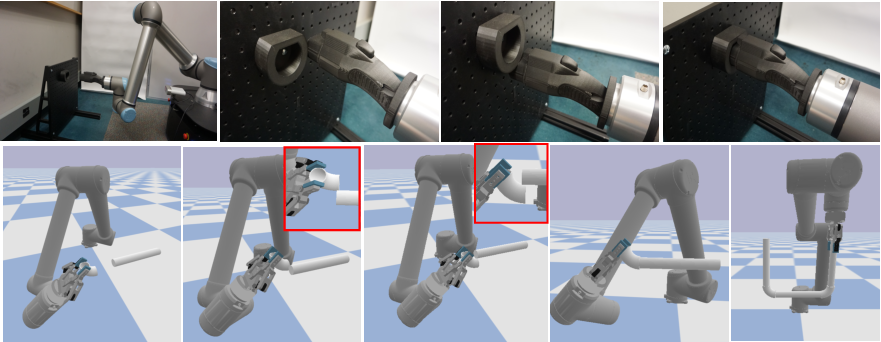}
  \caption{
  A run of the proposed framework on the plug insertion task in the real world (top row) and the pipe assembly task in simulation (bottom row). We observe the robot making contacts at different locations to localize the object of interest and completing the insertion task upon localization. In the pipe assembly case, the construction of complex structures can be modeled as a sequence of insertions.
}\label{fig:robot_experiments}\vspace{-0.3cm}
\end{figure*}





\begin{table}[]
\begin{center}
\caption{Statistics For Database Construction (Plug Insertion)}
\label{tab:plug insertion database}
\begingroup
\setlength{\tabcolsep}{4pt}
\begin{tabular}{@{}lccccccc@{}}
\toprule
\multirow{4}{*}{\textbf{Metrics}} & \multicolumn{7}{c}{\textbf{Planning Algorithm}} \\
\cmidrule{2-8}
& \multicolumn{4}{c}{\textsc{RTDP-Bel}} & \multicolumn{3}{c}{\textsc{E-RTDP-Bel}} \\ 
\cmidrule(l){2-5} \cmidrule(l){6-8}
& $\varepsilon = 1$ & $\varepsilon = 2$ & $\varepsilon = 3$ & $\varepsilon = 5$ & $\varepsilon = 2$ & $\varepsilon = 3$ & $\varepsilon = 5$ \\
\midrule
\multirow{2}{*}{\shortstack{Success \\ Rate ($\%$)}} & \multirow{2}{*}{78.9} & \multirow{2}{*}{81.1} & \multirow{2}{*}{84.2} & \multirow{2}{*}{86.5} & \multirow{2}{*}{\textbf{100}} & \multirow{2}{*}{\textbf{100}}  & \multirow{2}{*}{\textbf{100}}\\\\
 \midrule
\multirow{2}{*}{\shortstack{Relative \\ Speedup}} & \multirow{2}{*}{1.0} & \multirow{2}{*}{1.22} & \multirow{2}{*}{1.56} & \multirow{2}{*}{2.32} & \multirow{2}{*}{113.24} & \multirow{2}{*}{283.13}  & \multirow{2}{*}{\textbf{551.97}}\\\\
 \midrule
\multirow{2}{*}{\shortstack{Relative \\ Expected Cost}} & \multirow{2}{*}{\textbf{1.0}} & \multirow{2}{*}{1.08} & \multirow{2}{*}{1.78} & \multirow{2}{*}{2.19} & \multirow{2}{*}{1.12} & \multirow{2}{*}{1.83}  & \multirow{2}{*}{2.46}\\\\
 \bottomrule
\end{tabular}\vspace{-0.4cm}
\endgroup
\end{center}
\end{table}

\resp{Table \ref{tab:plug insertion database}} presents results that demonstrates the impact of the developed E-RTDP-Bel algorithm in constructing the database of policies $\Pi$. Each algorithm is given a timeout of 500 seconds for each planning problem. If the algorithm fails to solve the problem within the timeout, a failure is recorded. 
The speedups and costs are presented relative to RTDP-Bel (run with a suboptimality bound of $\varepsilon = 1$). From the table, it is clear that E-RTDP-Bel dominates RTDP-Bel providing average speedups of over a factor of 100 while maintaining similar costs. The table also presents the tradeoff between solution quality and planning time dictated by the penalty $\varepsilon$. As $\varepsilon$ increases, E-RTDP-Bel encourages the search to go farther out of its way and reuse larger portions of previous solutions thereby converging quicker at the expense of solution quality. The performance of RTDP-Bel under different suboptimality bounds highlights the relative impact of E-RTDP-Bel as 
the suboptimality bound increases. 

\subsubsection{Performance of the Overall Framework}

\begin{table}[]
\begin{center}
\caption{Performance of the Framework on plugin and assembly tasks}
\label{tab:real world results}
\begingroup
\setlength{\tabcolsep}{4pt}
\begin{tabular}{@{}lcc@{}}
\toprule
\multirow{2}{*}{\textbf{Metrics}} & \multicolumn{2}{c}{\textbf{Task}} \\
\cmidrule{2-3}
& \textsc{Plug Insertion} & \textsc{Pipe Assembly} \\
& \textsc{(Real World)} & \textsc{(Simulation)} \\
\midrule
\shortstack{Success Rate ($\%$)} & 95 & 100 \\ \midrule
\shortstack{Execution Time (\si{s})} & 22.3 & 16.9 \\ \midrule

\end{tabular}\vspace{-0.3cm}
\endgroup
\end{center}
\end{table}

Table \ref{tab:real world results} presents the performance of our framework averaged over 50 runs in the real world. In each run, the port is moved to a random pose within its workspace. We observe a strong success rate of 95\%. The source of the 5 failures is incorrect perception, wherein, the groundtruth pose is not contained within the hypothesis set identified by perception. 

\begin{figure}
  \centering
  \begin{minipage}[b]{0.49\linewidth}
    \includegraphics[width=\textwidth]{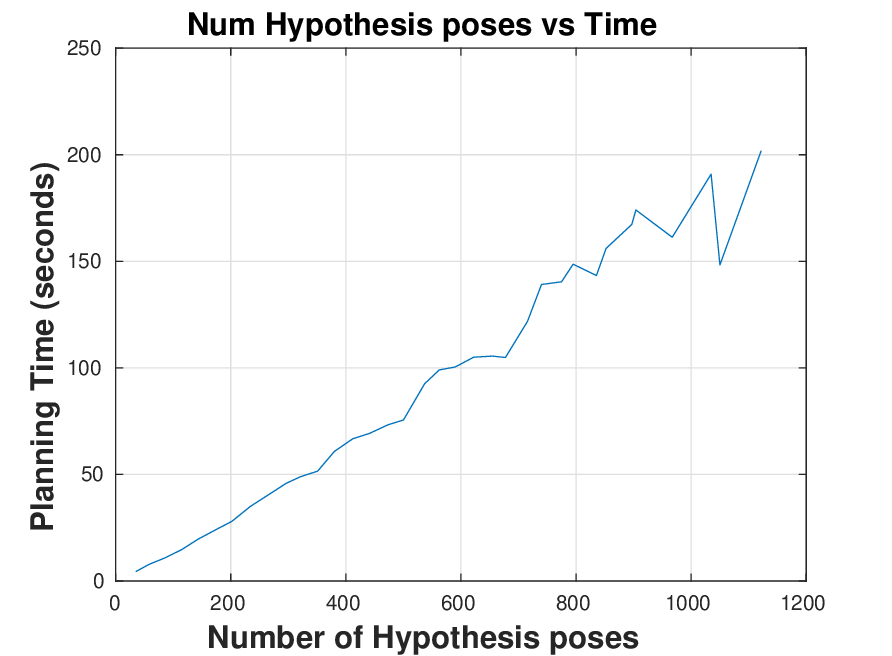}
  \end{minipage}
  \hfill
  \begin{minipage}[b]{0.49\linewidth}
    \includegraphics[width=\textwidth]{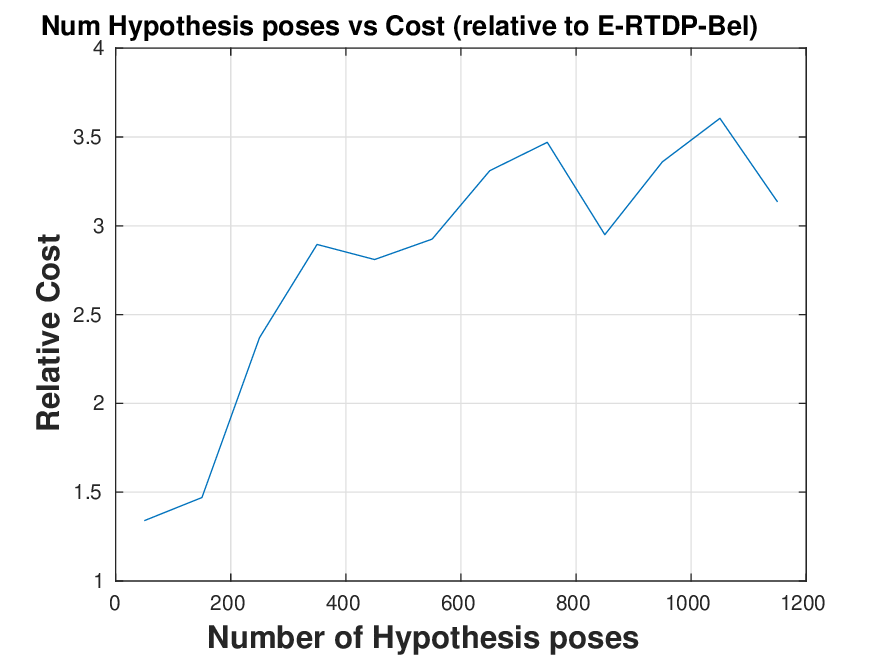}
  \end{minipage}
    \caption{Performance of TBL on the task of plug insertion. left) Initial uncertainty (number of hypothesis poses) vs planning time. right) Initial uncertainty (number of hypothesis poses) vs Cost (relative to RTDP-Bel). }
\label{Fig: TBL} \vspace{-0.5cm}
\end{figure}

We also evaluate a popular online approach, Touch Based Localization (TBL) \cite{TBL}, on our problem domain. TBL interleaves planning and execution. In each iteration of planning, multiple actions are sampled, and based on the current belief their expected information gain is computed. The action that maximizes the information gain metric is chosen for execution. Upon execution, the observation received is used to update the belief, and the process is repeated. This approach is inefficient in our domain for two reasons. The first is the computational expense associated with it, especially under high uncertainties. Evaluating information gain of an action or even maintaining the belief distribution, involves forward simulating the action under all hypothesis poses and computing the expected observation. This process involves expensive 3D mesh collision checks. In our problem, hypothesis sets can contain thousands of poses resulting in several thousand collision checks for each action. Although TBL is faster than solving the problem as a POMDP, it is not fast enough to be used online. The second reason is that TBL is myopic. It only reasons about the next best action to execute and not the sequence of actions to execute to complete the task, thereby compromising solution quality. This effect is exacerbated under high uncertainties. Fig \ref{Fig: TBL}. presents how the planning times and solution qualities scale with uncertainty. For our evaluation, we sampled 100 actions in the vicinity of the hypothesis poses in each planning iteration.

\resp{A simple Image-based visual servoing approach was also evaluated. A RealSense D435i camera was mounted on the wrist of the manipulator and provided a continuous video stream of the region of interest. SIFT \cite{resp-sift} was used for registering the features between the current and the target image. The positional errors between the features in the image space (from the current to the target image) were used to compute the robot controls. The lack of clear contrast between the object of interest and its background (black colored port mounted on a black background) combined with the fact that the plug in hand occluded most of the port (especially in close proximity) led to poor feature registration and as a result only 1 success out of 10 runs. To compare against the performance of the approach under ideal conditions, a clearly visible red tape that could be reliably identified in the images (including those from close proximity) was attached to the port. Instead of using the SIFT features, we used the corners of the tape (which could be reliably located using a simple color mask) as features for the IBVS algorithm. The performance was significantly better under such ideal conditions succeeding in 9 out of 10 runs (with an average execution time of 11 seconds). These results reiterate the challenges faced by vision-based algorithms and highlight the need for using tactile feedback.}

\begin{table}[]
\begin{center}
\caption{Preprocessing Framework Ablation Study}
\label{tab:ablation}
\begingroup
\setlength{\tabcolsep}{4pt}
\begin{tabular}{@{}lccc@{}}
\toprule
\multirow{3}{*}{\textbf{Metrics}} & \multicolumn{3}{c}{\textbf{Planning Framework}} \\
\cmidrule{2-4}
& \textsc{Ours} & \textsc{Naive} & \textsc{Random} \\
\midrule
\shortstack{Success Rate ($\%$)} & \textbf{100} & \textbf{100} & 95.5 \\ 
\midrule
\shortstack{Relative Speedup} & \textbf{1.0} & 0.23 & 0.11 \\
 \midrule
\shortstack{Relative Expected Cost} & 1.0 & \textbf{0.95} & 0.98 \\
 \bottomrule
\end{tabular}\vspace{-0.5cm}
\endgroup
\end{center}
\end{table}

We highlight the impact of the choices we make in our preprocessing-based planning framework through ablations presented in Table \ref{tab:ablation}. Here, $\textsc{Naive}$ uses $\mathcal{B}^E_{naive}$ as the experience MDP instead of the proposed rollout procedure to construct $\mathcal{B}^E$ (Section \ref{subsection: Using E-RTDP-Bel for Preprocessing}). $\textsc{Random}$ orders the problems randomly to highlight the benefit of ordering problems in increasing order of uncertainty. All variants run E-RTDP-Bel with $\varepsilon = 2$, and their performance in constructing the database of solutions is presented relative to our framework. We observe that $\textsc{Naive}$ and $\textsc{Random}$ are 5 and 10 times slower than our framework respectively, highlighting the impact of the \textit{quality} of experience on planning performance. By ordering the problems in terms of increasing uncertainty and by performing the rollout procedure, we are able to construct Experience MDPs that are more relevant (larger portions of the experience can be reused) and easily accessible, resulting in stronger performance.

Overall, our framework provides a strong combination of real-world robustness, planning times, and solution quality.

\subsection{Pipe Assembly Task}

\begin{table}[]
\begin{center}
\caption{Statistics For Database Construction (Pipe Assembly)}
\label{tab:pipe assembly database}
\begingroup
\setlength{\tabcolsep}{4pt}
\begin{tabular}{@{}lccccccc@{}}
\toprule
\multirow{4}{*}{\textbf{Metrics}} & \multicolumn{7}{c}{\textbf{Planning Algorithm}} \\
\cmidrule{2-8}
& \multicolumn{4}{c}{\textsc{RTDP-Bel}} & \multicolumn{3}{c}{\textsc{E-RTDP-Bel}} \\ 
\cmidrule(l){2-5} \cmidrule(l){6-8}
& $\varepsilon = 1$ & $\varepsilon = 2$ & $\varepsilon = 3$ & $\varepsilon = 5$ & $\varepsilon = 2$ & $\varepsilon = 3$ & $\varepsilon = 5$ \\
\midrule
\multirow{2}{*}{\shortstack{Success \\ Rate ($\%$)}} & \multirow{2}{*}{71.4} & \multirow{2}{*}{76.5} & \multirow{2}{*}{79.3} & \multirow{2}{*}{81.9} & \multirow{2}{*}{\textbf{100}} & \multirow{2}{*}{\textbf{100}}  & \multirow{2}{*}{\textbf{100}}\\\\
 \midrule
\multirow{2}{*}{\shortstack{Relative \\ Speedup}} & \multirow{2}{*}{1.0} & \multirow{2}{*}{1.4} & \multirow{2}{*}{1.76} & \multirow{2}{*}{2.06} & \multirow{2}{*}{84.44} & \multirow{2}{*}{198.10}  & \multirow{2}{*}{\textbf{335.50}}\\\\
 \midrule
\multirow{2}{*}{\shortstack{Relative \\ Expected Cost}} & \multirow{2}{*}{\textbf{1.0}} & \multirow{2}{*}{1.06} & \multirow{2}{*}{1.57} & \multirow{2}{*}{1.76} & \multirow{2}{*}{1.08} & \multirow{2}{*}{1.68}  & \multirow{2}{*}{1.89}\\\\
 \bottomrule
\end{tabular}
\endgroup
\end{center} \vspace{-0.7cm}
\end{table}

To demonstrate the generality of the framework, the approach was also evaluated in simulation on the task of pipe assembly. Pipe assembly can be modeled as a series of high-precision insertions of pipes into elbows and vice versa (Fig \ref{fig:robot_experiments}). For each insertion in this domain, we reason about the uncertainty in pose along 4 axes (x, y, yaw, and pitch) in contrast to just the positional uncertainty in plug insertion. 

As the task is evaluated in simulation, the set of initial uncertainties is artificially created and E-RTDP-Bel is used to create the database of policies. The set contains initial uncertainties ranging up to 2cm in position (along each axis) and 0.25 radians in orientation (along each axis). We used a resolution of 2mm for position and 0.05 radian for orientation for discretizing the distribution. Therefore, $\mathcal{H}$ has hypotheses sets $H_i$ that contain up to 3600 possible poses. During evaluation, random noise is added to the pose of the pipe/elbow. An initial distribution is sampled from the set $\mathcal{H}$ (such that the random noise is contained within the boundary of the discrete distribution), and the corresponding policy is executed by the robot. An important observation is that the random noise is sampled in the continuous space and is not exactly contained in the discrete distribution. The approach was evaluated on 50 problems and we observed that the robot was able to successfully localize the object of interest and complete the task in all of the runs (Table \ref{tab:real world results}). Similar to the plug insertion task, Table \ref{tab:pipe assembly database} presents the performance of E-RTDP-Bel in constructing the database of policies $\Pi$. E-RTDP-Bel continues to provide significant speedups while maintaining good solution quality. The ablation studies and performance comparisons with TBL follow similar trends as in the plug insertion task and are omitted for brevity. 

\section{Conclusion}
We formulate high-precision insertion tasks as planning under pose uncertainty and utilize contacts to localize the object of interest and complete the task. We develop a preprocessing-based planning framework for semi-structured insertion domains where the set of possible pose uncertainties are finite and known ahead of time. Due to the computational expense of preprocessing a database of solutions, we propose a general experience-based POMDP solver, E-RTDP-Bel, that effectively utilizes solutions of similar planning problems to speed up planning queries and use it to speed up database construction by over a factor of 100. The framework demonstrates strong performance in terms of robustness, planning times, and solution quality on the tasks of plug insertion (real world) and pipe assembly (simulated). 

\bibliographystyle{plain}
\bibliography{root}

\end{document}